\documentclass{article}

\usepackage{epsfig}
\usepackage{subcaption}
\usepackage{calc}
\usepackage{amssymb}
\usepackage{amstext}
\usepackage{amsmath}
\usepackage{amsthm}
\usepackage{multicol}
\usepackage{pslatex}
\usepackage{apalike}
\usepackage{algorithm2e}
\usepackage[bottom]{footmisc}
\usepackage{mathrsfs}
\usepackage[nice]{nicefrac}
\usepackage{comment}
\usepackage{siunitx}
\usepackage{hyperref}
\usepackage[acronym]{glossaries}
\usepackage{caption}
\usepackage{subcaption}

\newcommand{\tdlgm}{\large{t}\normalsize{DLGM}}

\DeclareMathOperator{\val}{\mathbf{v}}
\DeclareMathOperator{\latent}{\mathbf{\xi}}
\DeclareMathOperator{\values}{\mathbf{V}}

\DeclareMathOperator{\state}{\mathbf{s}}
\DeclareMathOperator{\states}{\mathbf{S}}

\newacronym{tdlgm}{\tdlgm}{Time Deep Latent Gaussian Model}
\newacronym{dlgm}{DLGM}{Deep Latent Gaussian Model}
\newacronym{sdlgm}{DLGM}{Deep Latent Gaussian Model}
\newacronym{mlp}{MLP}{Multi-Layer Perceptron}
\newacronym{rnn}{RNN}{Recurrent Neural Network}
\newacronym{mse}{MSE}{Mean Squared Error}
\newacronym{lstm}{LSTM}{Long Short-Term Memory}
\newacronym{slstm}{LSTM}{Long Short-Term Memory}
\newacronym{gru}{GRU}{Gated Recurrent Unit}
\newacronym{vrnn}{VRNN}{Variational Recurrent Neural Network}
\newacronym{vae}{VAE}{Variational Auto-Encoder}
\newacronym{llm}{LLM}{Large Language Model}
\newacronym{timegan}{Time-GAN}{Time Generative Adversarial Network}
\newacronym{gan}{GAN}{Generative Adversarial Network}
\newacronym{stdlgm}{\footnotesize t\small DLGM}{Time Deep Latent Gaussian Model}

\newcommand*\samethanks[1][\value{footnote}]{\footnotemark[#1]}
\begin{document}
\title{ 
Approximate Probabilistic Inference for Time-Series Data
\\ {\small A Robust Latent Gaussian Model With Temporal Awareness} 
}
\author{Anton Johansson\thanks{Department of Mathematics and Computer Science, Karlstad University. Email: \{firstname (dot) lastname\}@kau.se}\and Arunselvan Ramaswamy\samethanks
}
\date{}

\maketitle

\abstract{The development of robust generative models for highly varied non-stationary time series data is a complex yet important problem.
Traditional models for time series data prediction, such as \gls*{slstm} are inefficient and generalize poorly as they cannot capture complex temporal relationships. In this paper, we present a probabilistic generative model that can be trained to capture temporal information, and that is robust to data errors. We call it \gls*{stdlgm}. Its novel architecture is inspired by \gls*{sdlgm}. Our model is trained to minimize a loss function based on the negative log loss. One contributing factor to \gls*{tdlgm} robustness is our regularizer, which accounts for data trends.
Experiments conducted show that \gls*{stdlgm} is able to reconstruct and generate complex time series data, and that it is robust against to noise and faulty data.

\section{\uppercase{Introduction}}
\label{sec:introduction}

Time series prediction constitutes an important class of problems in the field of machine learning.
It finds applications in numerous areas, such as traffic and demand prediction in the fifth generation of communication networks (5G), weather forecasting, traffic prediction in vehicular networks, etc. 
In such scenarios, the arising time series data is highly varied and noisy, and the associated data distributions are non-stationary (time-dependent).
Additionally, collecting this data is often expensive and time-consuming.
One important example is in the field of wireless communication, e.g., 5G, where researchers collect customer usage patterns and traffic, and performance of service providers, over time.
Such data, in addition to being expensive is often not publicly available \cite{5gdata}.
Another scenario with the aforementioned characteristics is the field of self-driving cars\cite{car}.
Both the 5G and self-driving car scenarios contain highly varied time-series data and, as such, exhibit complex temporal patterns.
The traffic load in a 5G network is highly dependent on the time of day.
Likewise, the traffic behavior in a vehicular network varies based on region, type of road, time of day, etc.
There is a need to develop robust, probabilistic models for time-series data prediction. 
Such models can be used for robust planning, augmenting small datasets (robust synthetic data generation), value imputation to overcome noise, training AI agents in a robust manner, etc., for a variety of problems that arise in the above mentioned example scenarios.
Robustness is an important property since the available data is often noisy.
In this paper, we are additionally interested in probabilistic models since they are expected to generalize better.
\if
  One common denominator between all these fields is the need for large amounts of data.
Some problems require tens of thousands, millions, or even sometimes billions of data points to train.
This is not an issue in some domains where it is widely available.
\glspl*{llm}, for example, is one kind of model that trains on textual data, something which is widely available online \cite{llama}.
Companies that develop \glspl*{llm} benefit from their large budgets and easily accessible training data.
However, there are other fields where data is less available.
For example, in networking and 5G/B5G, where the data can be costly and time-consuming to collect \cite{kau}, and there is a lack of public data \cite{5gdata}.
Or in the case of self-driven vehicles where most, if not all, data has to be manually collected by driving \cite{car}.
Both the 5G/B5G and self-driven scenarios contain highly varied time-series data and, as such, exhibit complex temporal patterns.
For example, the traffic load on a 5G network can depend highly on the time of day.
Likewise, traffic behavior varies based on region, type of road, etc.
The limited datasets combined with the complex, often non-stationary data require a robust time series model.  
\fi

A \glspl*{rnn} is a simple and popular model for time series data prediction. It varies from regular feed-forward networks through the use of specialized time-aware neurons. \gls*{lstm} and \gls*{gru} are two important choices for neurons when building \glspl*{rnn} \cite{lstm,gru}. Our model, \gls*{tdlgm}, is built using \gls*{lstm} units. Every \gls*{lstm} unit has a cell state, a value that aims to capture short-term as well as long-term temporal information passing through that unit. Let $\state$ represent the vector of all the \gls*{lstm} cell states from an \gls*{rnn}, and let $x$ be the input to the \gls*{rnn}. Passing $x$ through the said \gls*{rnn} changes the cell states of all the constituent \gls*{lstm} units to $\state'$. For the sake of clarity in presentation, we abstractize this operation using a function $F: \{\text{set of all possible cell states}\} \to \{\text{set of all possible cell states}\},$ with $\state' = F(\state, x).$

\glspl*{rnn} are prone to overfitting, hence not robust to errors in data collection. 
Further, they perform poorly when the dataset is highly varied and complex. 
Our model overcomes these issues by using the above-described state information $\state$ in order to predict the next-step data. 
In other words, \gls*{tdlgm} uses latent relevant information from the recent past in order to predict the next data point. 
In particular, it predicts the parameters of the probability distribution of the next data point. 
There are other stochastic models for time-series data prediction, e.g., \gls*{timegan} and \gls*{vrnn}  \cite{timegan,vrnn}.

\subsection{Literature Survey}

Previous nondeterministic models have been developed to address the issue of complex time series data.
Two notable examples are \gls*{timegan} and \gls*{vrnn} \cite{timegan,vrnn}.
Both models share a common characteristic, which is the idea of modeling a latent variable.
We define $\latent$ as a vector of latent variables and $\val\in \values$ as values from a time series dataset.
These models are then based on the idea of \gls*{vae}, which is used in situations where the prior of a latent variable is known ($p(\latent)$), but the posterior ($p(\latent|\val)$) is not \cite{vae}.
If the relationship between $\latent$ and $\val$ were discovered, new data points could be accurately generated by sampling from $p(\latent)$.
\gls*{vae} address this unknown relationship by approximating posteriors ($p(\latent|\val)$) through a recognition model.
The approximated posteriors are then used to train a generator model.
The generator model can then create new data points by sampling from the prior.
\gls*{vrnn} and \gls*{timegan} do this but with the additional constraint that latent variables are conditioned on the state.

\gls*{timegan} is based on the idea of a \gls*{gan} \cite{timegan,gan}.
That is a type of model with one generator and one discriminator. 
The generator is trained to create values, while the discriminator is tasked with discerning true and fake values.
\gls*{timegan} moves this to the latent space, meaning that the discriminator tries to discern between true and generated latent variables $\latent$.
While the generator is tasked with fooling the discriminator.
The latent variables are parallel to this used to train another model, which constructs $\val$ from $\latent$.

\gls*{vrnn} is a more straight forward usage of inference \cite{vrnn}.
It trains a set of neural networks based on previous states that approximate a distribution for the latent variable.
Samples from this distribution are then used to create a value.
Our model has some properties similar to \gls*{vrnn}.
We will, therefore, discuss \gls*{vrnn} in further detail in the next section.

\subsection{Our Contributions and Place in Literature}

As previously stated, \gls*{vrnn} is based on the idea of \gls*{vae}, which is used in situations where the prior of a latent variable is known ($p(\latent)$), but the posterior ($p(\latent|\val)$) is not \cite{vae}.
\gls*{vrnn} solves this by training a function that extracts latent variables from previous states.
\gls*{vrnn} does this through two samples per time-step $t$.
Specifically, given previous state $\state_{t-1}$ they define a latent variable as 
\begin{equation}
\latent_t \sim \mathcal{N}(\mu_{0,t},\sigma^2_{0,t}), \text{ where, }[\mu_{0,t},\sigma_{0,t}] = p(\state_{t-1})
\end{equation}
where p is a trainable function.
Hence, the previous state defines a Gaussian distribution.
This is then used to sample a value $\val$
\begin{equation}
\val_t \sim \mathcal{N}(\mu_{x,t}, \sigma^2{x,t}), \text{ where, } [\mu_{x,t},\sigma_{x,t}] = p_x(p_z(\latent_t),\state_{t-1})
\end{equation}
$p_x$ and $p_z$ are both trainable functions.
This structure, with one sample for the latent variable and another to generate $\val$, works well for time series data.
However, we believe that two layers of sampling hinder the potential robustness of the generative model.
More samples can result in more intricate distributions.
Therefore, we want a generative model for time series data in which the layers of combined samples can be set as a parameter of the model.

\gls*{dlgm} was developed by Rezende et al. in 2014 to solve the issue of scaleable inference in deep neural network model \cite{dlgm}.
It is constructed by performing approximate inference in layers and, as such, combining multiple Gaussian samples.
This means that the number of layered samples can vary depending on the dataset's needs.
Resulting in the possibility for more complex distributions, as compared to \gls*{vrnn} where there are always two layers.
This allows the model to learn complex patterns, generate new values, and perform inference.
However, it cannot, despite these excellent properties accommodate time series data.
We address this by combining \gls*{dlgm} with the idea of conditioning latent variables on previous states.
The result is a novel recognition-generator structure that utilizes two recognition models, one for state and one for latent variables. 
It differs from \gls*{vrnn} through the use of two recognition models and the interleaving of state and latent variables.
To our knowledge, this model, called \gls*{tdlgm}, is new in the literature.
We believe this interleaving of state and latent variables structure should be as good or better than \gls*{vrnn} at learning temporal information from a dataset. 
The structure of sampling in layers should allow it to learn more complex distributions and be more robust against erroneous or faulty data.
Our belief is based on the fact that a single-layered \gls*{tdlgm} reduces down to a structure similar to that of \gls*{vrnn}.

This paper is a continuation of the master thesis of Anton Johansson \cite{masterthesis}.
This paper further develops the work presented there, with an added discussion about robustness and improvements in the derivation of our loss function.

The rest of this paper is organized as follows.
First comes a section that presents the main equations of \gls*{dlgm} and the modifications done to construct \gls*{tdlgm}.
Following this is a series of derivations resulting in a well-defined generative model.
Our third chapter presents the experiments, the results of which are discussed in chapter four.
We then end with a conclusion chapter summarizing the paper and discussing future directions.

\section{Time-Series Deep Latent Gaussian Model}

Let us start with a few definitions.
We have a time series dataset called $\values$. An entry from this dataset is called $\val$ where $\val_t$ is the value generated at time $t$.
$\val_t$ is then followed by $\val_{t+1}$ \textit{ad infinitum}, we assume that $t \in \mathtt{N}$, that is $t$ can take all of the natural numbers.
All values $\val \in \values$ are assumed to be generated through an unknown probabilistic function. 
Hence, we have a probability function $p(\val)$ that needs to be defined.
We define this based on the idea of latent variables. 
Such that every $\val$ can be correlated to a latent variable $\latent$.
These latent variables are drawn from the Gaussian distribution $\mathcal{N}(\mathbf{0},\mathbf{I})$, where $\mathbf{0}$ is the zero vector and $\mathbf{I}$ is the identity matrix.
The decision of the distribution function is itself entirely arbitrary.
However, choosing a \textit{nice} Gaussian simplifies the loss function.
This means the prior $p(\latent)$ is known, but the posterior $p(\latent | \val)$ is unknown.
Approximating this posterior allows for training a generator model that can create new data points \cite{vae}.
The approximation function is often called the recognition model, resulting in a recognition-generator model pair.
This is how \gls*{dlgm} and similar methods structure their model \cite{dlgm}.
Our model extends this by introducing the idea of a state called $\state$. 
A value $\val$ at time $T$ is then defined as dependent on all previous states and latent variables.
\begin{equation}\label{eq:next_state}
    p(\mathbf{s}_{T}) = p(\mathbf{s}_{0})\prod_{t=1}^{T-1}\left[p(\mathbf{s}_{t}| \mathbf{s}_{1:t-1}, \mathbf{\xi}_{t})\right],
\end{equation}
where $\state_{1:t-1} = (\state_1,\ldots,\state_{t-1})$.
Our state definition depends on the structure of \gls*{tdlgm} and will be defined more rigorously later.

The idea behind our proposed model \gls*{tdlgm} originates from the structure of \gls*{dlgm} \cite{dlgm}.
It is based on the idea that the sampled values $\val$ are constructed from latent variables $\latent$. It is, therefore, possible, given that the posterior $p(\mathbf{\xi}|\mathbf{v})$ is known to train a model capable of creating new values based on latent samples. 

\gls*{dlgm} use independently sampled Gaussian variables ($\mathbf{\xi})$ in combination with \glspl*{mlp} $T$ on each layer to form an interleaving structure.
\gls*{dlgm}'s generator model can be defined through a set of equations:
\begin{equation}
    \mathbf{\xi}_l \sim \mathcal{N}(\mathbf{\xi}_l | \mathbf{0}, \mathbf{I}), \quad l = 1,\ \ldots, \ L
\end{equation}
\begin{equation}
    \mathbf{h}_L = \mathbf{G}_L\latent_L,
\end{equation}
\begin{equation} \label{dlgm_h}
    \mathbf{h}_l = T_l(\mathbf{h}_{l+1}) + \mathbf{G}_l\mathbf{\xi}_l, \quad l=1, \ \ldots, \ L-1
\end{equation}
\begin{equation}
    \mathbf{v} \sim \pi(\mathbf{v}|T_0(\mathbf{h}_1)).
\end{equation}
In the above set of equations, $\mathbf{G}_l$ is defined as trainable matrices, and $T_l$ are \glspl*{mlp}.
The posterior is, as previously discussed, unknown.
This is addressed by approximating $\latent$ on each layer through a recognition model $q(\latent | v) \approx p(\latent | v)$.
The generator model is then trained to recreate $\val$ based on values provided by the recognition model.
Note the lack of any mention of state in the above generator.
This is because \gls*{dlgm} is a stateless model, which we will now address.


\gls*{tdlgm} is obtained by modifying \gls*{dlgm} by replacing \glspl*{mlp} in \eqref{dlgm_h} with \glspl*{rnn}.
Its architecture is illustrated in Figure \ref{fig:tdlgm}.
A key ingredient in the design of \gls*{tdlgm} is the concept of a "latent state" that we simply refer to as a state. 
We previously postponed the rigorous definition of a state. 
Now is the time for that.
A state $\state$ at time $t$ called $\state_t$ is a vector that stores relevant temporal information. 
This $\state_t$ structure depends on the internals of the used time series model.
One common characteristic of a state vector is its finite size, which prevents it from storing all previous information.
Hence, there must be a transition function $F$ from $\state_t$ to $\state_{t+1}$ that decides what to keep from one state to the next.
Where the definition of $F$ depends on the model's design.
One naive method is to structure $\state$ as a queue of previous results with a set length $n$.
That is $F(\val_{t-1}, \state_{t-1}) = \state_t = [\val_{t-1},\  \ldots,\ \val_{t-1-n}]$.
However, the more naive implementations of a state transition function tend to have issues with exploding or vanishing gradients \cite{lstm}.
We will base our state representation on the previously discussed \gls*{lstm} as it is proven to be more resistant to these problems.
\gls*{lstm}, as previously discussed, contain something called a cell state.
This cell state intends to store \textit{relevant} temporal information.
$F$ is trained through gradient descent to \textit{learn} what should or should not be stored.
The model uses this to \textit{remember} important information and \textit{forgetting} that which is not needed.
Recall Equation \eqref{eq:next_state} that the state $\state_t$ depends on all previous latent variables and states.
This is no longer possible given the constraints that all $\state \in \states$ have the same size.
Based on this, we introduce an upper limit $m$, which is the point beyond which past states and latent variables have no discernible impact. 
Hence,
\begin{equation}\label{state_depend}
    p(\state_T) = \prod_{t=T-m}^{T-1}[p(\state_t | \state_{T-m:t-1}, \latent_{t-1})],
\end{equation}
where $\state_{T-m:t-1} = [\state_{T-m},\ \ldots \ \state_{t-1}]$
This transition can be seen on the left side of Figure \ref{fig:tdlgm}, where each layer has a state and corresponding transition function.
One idea which remains the same between \gls*{tdlgm} and \gls*{dlgm} is the notion of layers.
That is, both use the output of one layer ($h_l$) as the input to the next ($h_{l-1}$) until the bottom layer is reached.

\glspl*{tdlgm} generator model can therefore be defined as:
\begin{equation}\label{sample}
    \mathbf{\xi}_{l,t} \sim \mathcal{N}(\mathbf{\xi}_l | \mathbf{0}, \mathbf{I}), \quad l = 1 \ ,\ldots, \ L
\end{equation}
\begin{equation}\label{hL}
    \mathbf{h}_{L,t} = \mathbf{G}_L\mathbf{\xi}_{L,t},
\end{equation}\label{hl}
\begin{equation} \label{tdlgm_h}
    \mathbf{h}_{l,t} = R_l(\mathbf{h}_{l+1,t},\mathbf{s}_{l,t}) + \mathbf{G}_l\mathbf{\xi}_{l,t}, \quad l=1, \ \ldots, \ L-1
\end{equation}
\begin{equation}\label{next_state}
    \mathbf{s}_{l,t+1} = F_l(\mathbf{h}_{l-1,t},\mathbf{s}_{l,t}), \quad l = 1,\ldots,L-1
\end{equation}
\begin{equation}\label{result}
    \mathbf{v}_t \sim \pi(\mathbf{v}|T_0(\mathbf{h}_{1,t})),
\end{equation}
each of the layers $l=1,\ldots,L-1$ has its own \gls*{rnn} $R_l$ and state $\state_{l,t}$, where $t$ denotes the step in time.
We use \gls*{lstm} for our \gls*{rnn}; however, the design remains the same for other types of neurons.
$\state_{2,5}$ is therefore the state at layer $2$ at timestep $5$.
$F_l$ on each layer is the state transition function which is dependent on its corresponding \gls*{rnn}.
Therefore, the chosen \gls*{rnn} dictates the nature of the state $\state$ and state transition function $F$.
Other components of \gls*{tdlgm} remain similar to \gls*{dlgm}.
We will now go through the whole generation process:
The model starts with a sample $\latent_{l,t}$ from a Gaussian with the zero vector as mean and identity matrix as covariance.
It is combined with the matrix $\mathbf{G}_L$, resulting in $H_L$.
$H_L$ is then used in the layer below as input to the \gls*{rnn} $R_{L-1}(H_{L}, \state_{L-1,t})$.
The sum of $R_{L-1}$ and $G_{L-1}\state_{L-1,t}$ becomes the output of that layer, also called $h_{L-1}$, which is used in the layer below called $h_{L-2}$.
This repeats down to the last layer $h_1$, which is then used in Equation \eqref{result}.
The $h$ from the layer above ($h_{l+1}$) is also combined with $\state_{l,t}$ to create the next state as seen in Equation \eqref{next_state}.
The layered latent variables provide robustness to the stateful model by interleaving states and latent variables.

\begin{figure}
    \centering
    \includegraphics[width=0.5\textwidth]{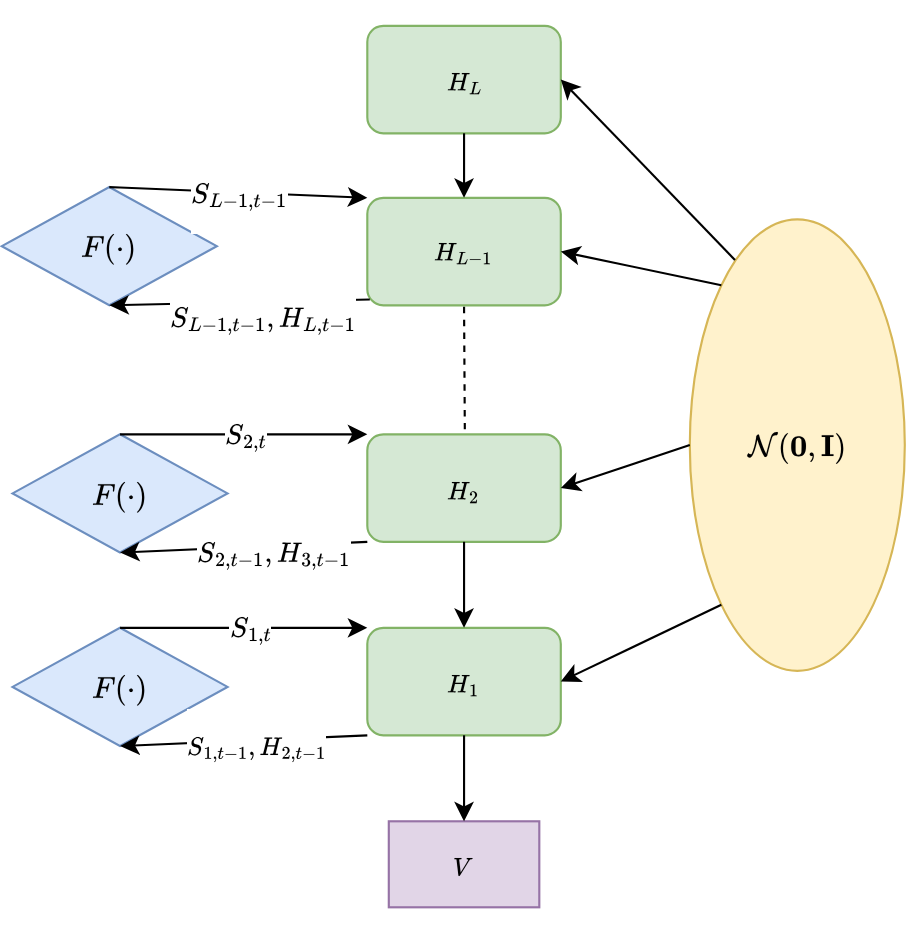}
    \caption{
    This figure shows how values are generated with \gls*{tdlgm}. It start at $H_L$ which calculates its value by sampling from $\mathcal{N}\mathbf{0},\mathbf{I})$.
    This result is then passed down to $\mathbf{H}_{l-1}$, which utilizes $H_L$, the current state, and a sample to calculate its value.
    This is done until the end, resulting in $v$ as specified in \eqref{result}.
    It is also shown on the left-hand side how the state is updated by the function $F$ on each layer.
    }
    \label{fig:tdlgm}
\end{figure}

Recall that the posterior was unknown for \gls*{dlgm} and, therefore, had to be approximated.
The same issue is present for \gls*{tdlgm} as we need to know both $p(\state | \val)$ and $p(\latent | \val)$.
This is solved by introducing two recognition models, one being the state recognition model defined as $k$ and another being a recognition model for the latent variable defined as $q$.

Starting with $q$, which can utilize a method similar to that of \gls*{dlgm}.
That is, we define a posterior for the latent variable as
\begin{equation}\label{latentpost}
    q(\latent|\values) = \prod_{t=1}^T\prod_{l=1}^L \mathcal{N}(\latent_{l,t}|\mu_l(\val_t),\mathbf{C}_l(\val_t)),
\end{equation}
where the mean $\mathbf{\mu}$ and covariance $\mathbf{C}$ on each layer are trainable parameters.
We take another approach to the state recognition model.
Recall Equation \eqref{next_state}, which is responsible for creating future states that are, as defined by Equation \eqref{state_depend}, dependent on $m$ previous states and latent variables.
If the $m$ previous states and latent variables were known, then it would be possible to train $F$ directly.
However, since this is not known. We instead have to approximate the state on each layer
\begin{equation}
    F_l( \state_{l,t},\mathbf{h}_{l+1:L,t}) \approx \hat{F}_l( \mathbf{v}_{t-m:t}) = \state_{l,t}
\end{equation}
where $l$ is the corresponding layer, see Equation \eqref{next_state}.
Each layer has one $F$ and $\hat{F}$.
We will omit this going forward and write $\hat{F} \approx F$.
Where $F = [F_1(\state_{t,1},h_{t,l+1}),\ldots,F_{L-1}(\state_{t,L-1},h_{t,L})]$  and $\hat{F} = [\hat{F}_1(\val_{t-m:t}),\ldots,\hat{F}_{L-1}(\val_{t-m:t})]$
A slight abuse of notation allows us to define the function $F$ and $\hat{F}$ in terms of $\state$ and $\latent$ instead of using $\state$ and $h$.
This is possible as $h$ depends entirely on $\state$ and $\latent$, as seen in Equations \eqref{hL} and \eqref{hl}.
It is done because our redefinition allows us to arrive at a cleaner-looking loss function without altering the underlying implementation.
We can then define the approximation as
\begin{equation}
    k(\states|\values) = \prod_{t=1}^T\prod_{l=1}^{L-1}p(\state_{l,t} | \hat{F}_{l}(\val_{t-m:t-1})),
\end{equation}
or stated another way
\begin{equation}
    p(\state_t|\state_{t-m:t-1}, \latent_{t-m:t-1}) \approx k(\state_t|\val_{t-m:t-1}),
\end{equation}
this solution is not yet complete.
Readers may have already noticed that the prior $p(\state)$ is unknown and that $F$ and $\hat{F}$ are deterministic functions.
This means that the methods used for the latent variable recognition model will not work.
This will be solved by deriving our loss function.

\subsection{Deriving the loss function}
We are now ready to derive the loss function that the \gls*{tdlgm} must minimize in order to accomplish its objective. Broadly speaking, the goal of \gls*{tdlgm} is to maximize the likelihood of generating the training data $\mathbf{V}$. Technically, it is achieved by minimizing the negative log-likelihood loss function given by:
\begin{equation}
    \begin{split}
    \mathcal{L}(\mathbf{V}) &= -\log p(\mathbf{V}) 
    \\ &= -log \int_{\mathbf{S}}\int_{\mathbf{\xi}}p(\mathbf{V}|\mathbf{\xi},\mathbf{S})p(\mathbf{\xi},\mathbf{S})d\mathbf{\xi}d\mathbf{S}
    \end{split}
\end{equation}
our negative likelihood function is an integration over $\states$ and $\latent$.
This assumes that $\states$ and $\latent$ are independent. 
We argue that this is the case due to Equation \eqref{state_depend}, which says that only the $m$ latest values influence the next state.
Both $\state$ have a finite size, while the generator can create infinite values. 
You cannot fit an infinite amount of previous states in a value $\state$ of finite size.
There must, therefore, be a repeat of state.
Further, $\latent$ is sampled independently of $\state$ for every generation, as defined in Equation \eqref{sample}.
Hence, given a number of generations such that $\state_k=\state_m, m\neq k$, then we can say that almost surely $\latent_k\neq\latent_m$.
This assumption ties back into our idea of robustness as we do not want the generator model to map any specific $\state$ to a $\val$.

Both $\state$ and $\latent$ have unknown posteriors $p(\mathbf{s}|\mathbf{v})$ and $p(\mathbf{\xi}|\mathbf{v})$ which is addressed in the previous chapter by introducing two recognition models: the state recognition model $k(\mathbf{s}|\mathbf{v}) \approx p(\mathbf{s}|\mathbf{v})$ and the latent recognition model $q(\mathbf{\xi}|\mathbf{v}) \approx p(\mathbf{\xi}|\mathbf{v})$
Both recognition models are included in the loss.
Jensen's inequality is then used to get a surrogate loss.
\begin{equation}
    \begin{split}
        \mathcal{L}(\mathbf{V}) &= -log \int_{\mathbf{S}}\int_{\mathbf{\xi}}\frac{k(\mathbf{s})q(\mathbf{\xi})}{k(\mathbf{s})q(\mathbf{\xi})}p(\mathbf{V}|\mathbf{\xi},\mathbf{S})p(\mathbf{\xi},\mathbf{S})d\mathbf{\xi}d\mathbf{S} \\
        &\leq -\int_{\mathbf{S}}\int_{\mathbf{\xi}}k(\mathbf{s})q(\mathbf{\xi})log\left(\frac{ p(\mathbf{V}|\mathbf{\xi},\mathbf{S})p(\mathbf{\xi},\mathbf{S})}{k(\mathbf{s})q(\mathbf{\xi})} \right)d\mathbf{\xi}d\mathbf{S} \\
        &\leq \int_{\mathbf{S}}\int_{\mathbf{\xi}}k(\mathbf{s}) q(\mathbf{\xi})log\left(\frac{k(\mathbf{s})}{p(\mathbf{s})} + \frac{q(\mathbf{\xi})}{p(\mathbf{\xi})} -  p(\mathbf{V}|\mathbf{\xi},\mathbf{S}) \right)d\mathbf{\xi}d\mathbf{s},
    \end{split}
\end{equation}
using the fact that the integral of a probability density function is equal to one leaves us with
\begin{equation}
\begin{split}
    \mathcal{L}(\mathbf{v}) \leq 
        D_{KL}(k(\mathbf{S})||p(\mathbf{S})) + 
    D_{KL}(q(\mathbf{\xi})||p(\mathbf{\xi})) - \\
    \mathbb{E}_{q(\mathbf{\xi}),k(\mathbf{S})}[
        log(
            p(\mathbf{V}|\mathbf{s},\mathbf{\xi})
        )
    ].
    \end{split}
\end{equation}

Recall the definition of the posterior in Equation \eqref{latentpost}.
We use this to say that
\begin{equation}\label{dklp}
    D_{KL}(q(\mathbf{\xi})||p(\mathbf{\xi})) = D_{KL}(\mathcal{N}(\mathbf{\mu},\mathbf{C})||\mathcal{N}(\mathbf{0},\mathbf{1})).
\end{equation}
which will later be solved analytically.

The probability for $k$ has been previously defined where:
\begin{equation}
    p(\mathbf{s}_t | \mathbf{s}_{t-m:t-1}, \mathbf{\xi}_{t-m:t-1}) \approx k(\mathbf{s}_t | \mathbf{v}_{t-m:t-1}),
\end{equation}
state, latent variables, and how they influence the current state is approximated based on true values $\mathbf{v}$.
It is then possible to use an approximated state for the generation of future states
\begin{equation}
    \begin{split}
    p(\state_{t+1} | F(\state_{t-m:t},\latent_{t-m:t})) 
        \approx p(\state_{t+1}| F(\hat{F}(\val_{t-m:t-1}), \latent_t)) \\
    \end{split}
\end{equation}
meaning that we use $\hat{F}$ to approximate an intermediate state $\state_t$.
This intermediate state is then used in the generator model to generate the next state $\state_{t+1}$.
This is useful as it allows us to generate an approximate state through two methods.
The first method uses the approximation function $\hat{F}$.
While the second method receives an approximation from $\hat{F}$, which is then used in the true state transition function $F$.
Applying this to the KL divergence results in
\begin{equation}
    \begin{split}
        D_{kl}(k(\mathbf{S})||p(\mathbf{S})) &=
        D_{kl}(k(\mathbf{S}_{t+1})||p(\mathbf{S}_{t+1}))  \\
        &\approx  D_{kl}(k(\mathbf{S}_{t+1})||p(F(k(\mathbf{S}_{t}| \mathbf{V}_{t-m:t}), \latent_t)).
    \end{split}
\end{equation}
Our final issue arises because it is difficult to calculate the above KL-divergence.
Because both $F$ and $\hat{F}$ are deterministic functions and the prior $p(\state)$ is unknown.
We addressed this by calculating the \gls*{mse} instead
\begin{equation}\label{mseloss}
    D_{kl}(k(\mathbf{S})||p(\mathbf{S})) \simeq \alpha\text{MSE}(\hat{F}(\mathbf{V}_{t+1}),F(\hat{F}(\mathbf{V}_t), \latent_t))
\end{equation}
$\alpha$ is added as a scaling factor to account for the fact that the KL-divergence and \gls*{mse} are different metrics.
To summarize, we approximate $\state_{t}$ with function $\hat{F}(\val_{t-m:t-1}) = \state_{t}$.
Our approximation is then used to generate $\state_{t+1}$, $F(\hat{F}(\val_{t-m:t-1}), \latent_t) = \state_{t+1} $.
This generated state is then compared against an approximated state through
$\text{MSE}(\hat{F}(\val_{t}),F(\hat{F}(\val_{t-1}, h_t)))$.

Equation \eqref{dklp} and \eqref{mseloss} can then be applied in the surrogate loss
\begin{equation}
    \begin{split}
        \mathcal{L}(\mathbf{V}) &\leq 
        D_{KL}(k(\mathbf{S})||p(\mathbf{S})) + 
        D_{KL}(q(\mathbf{\xi})||p(\mathbf{\xi})) \\ &\quad- 
    \mathbb{E}_{q(\mathbf{\xi}),k(\mathbf{S})}[
        log(
            p(\mathbf{V}|\mathbf{s},\mathbf{\xi})
        )
    ] \\
    &\approx 
       D_{KL}(\mathcal{N}(\mathbf{\mu},\mathbf{C}))||\mathcal{N}(\mathbf{0},\mathbf{I}))  \\ &\quad+ 
         \alpha\text{MSE}(F(\mathbf{V}_{t+1}),F(\hat{F}(\mathbf{V}_t),h(\latent_t) \\ &\quad- 
    \mathbb{E}_{q(\mathbf{\xi}),k(\mathbf{S})}[
        log(
            p(\mathbf{V}|\mathbf{s},\mathbf{\xi})
        )
    ],
    \end{split}
\end{equation}
the KL-divergence between two Gaussians can be calculated analytically, resulting in the final loss
\begin{equation}\label{eq:final}
\begin{split}
    \frac{1}{2}\sum_{l,n}[
    ||\mu_{n,l}||^2 &+ Tr(C_{n,l}) - \log{|C_{n,l}|} - 1] \\ &+
    \alpha~MSE(k(\pmb{S}^+),p(\pmb{S}^+)) \\ &- 
    \mathbb{E}_{q(\pmb{\xi}),k(\pmb{s})}[
        log(
            (
                p(\pmb{v}|\pmb{s},\pmb{\xi},\theta)
                p(\theta)
            )
    ]
\end{split}
\end{equation}



\section{\uppercase{Experiments}}

Our experiments were performed with a dataset provided by Alibaba under a free license for research use \cite{alibaba}.
It is a tabular dataset containing traces of different data center tasks with their timestamps, resource requirements, and more.
We use only one of these entries since we want a simple use case for our tests,
namely the GPU requirements of a given task called \texttt{GPU\_MILLI} in the dataset. 
Arrival, deletion, and schedule times are discarded, and data points are assumed to be at a constant time from each other.

\gls*{tdlgm} is compared to two baseline models. 
Our first baseline model is a \gls*{rnn} implemented with \gls*{lstm} \cite{lstm}.
The second baseline model implements \gls*{dlgm} \cite{dlgm}.
\gls*{dlgm}, as previously discussed, is stateless and, therefore, is expected to perform worse than \gls*{tdlgm} as it can only go back one step in time. Our dataset exhibits complex connections between data points that are separated by long horizons.
In order to still use it as a baseline for the current time series scenario, we provide as input a history of data points for the next-step prediction.
Thereby explicitly providing temporal information that would otherwise be missed by \gls*{dlgm}.
Note that the input's dimensionality is artificially increased to accommodate past information (data points).
\gls*{rnn} and \gls*{dlgm} are known to be non-robust.
Specifically, the authors of \gls*{dlgm} found that it could not reconstruct outside the dataset if it were not trained with noisy data \cite{dlgm}.
In order to boost their robustness, an artificial noise is added to the training dataset, a common trick in Machine Learning literature.
In what follows, we compare their performance to \gls*{tdlgm} that is trained on both the noisy and unaltered training dataset. \emph{In short, it beats the baselines.}

Our experiments explored the following three scenarios that are interesting use cases for \gls*{tdlgm}:
\begin{enumerate}
    \item \textbf{Imputation,} where we feed noisy data (additive Gaussian noise) into our models and see how well they reconstruct the true data.  
    The imputation is performed on test data that was not used during training. As the dataset is simple, and since we explicitly feed \gls*{dlgm} with temporal information, we expect it to perform well, if not better than \gls*{tdlgm}. \gls*{rnn}, on the other hand, has the worse performance, as illustrated in Figure~\ref{fig:rnn_best_recon}. We repeated the experiments with varying levels of added Gaussian noise.
    \item \textbf{Multiple time-step prediction,} where we require the models to predict over multiple time-steps in the future, given the current value. This set of experiments is important since it is known that prediction errors typically accumulate over time when predicting over multiple steps. This is because predicted values are typically used to predict more values further into the future.
    We may consider two metrics: the average mean squared error over the prediction horizon and the similarity of the generated values to the true distribution.
    We use the latter since we are interested in the model's ability to capture trends in the given time series dataset. We are not interested in exactly duplicating the dataset, we also want to generate new data points.
    Standard metrics such as \gls*{mse} score how well a generative model can predict given data points.
    One of our future goals is to use this generative model in cases where the data is limited.
    If it were to recreate the already limited dataset accurately, it would lose its purpose.
    We want it to generate new but similar data, not the same data.
    The used metric is explained in Algorithm \ref{alg:score}.
    \gls*{tdlgm} to performed best in this, followed by \gls*{dlgm} and then \gls*{rnn}.
    \if
    Models are, therefore, scored by grouping the generated values into intervals.
    These intervals are then scored based on how the frequency of future values correlates with the true future data.
    This allows us to score how similar the distribution is to the true distribution.
    Without penalizing the models for creating novel data within the expected structure, as would be the case if we used \gls*{mse}.
    Algorithm \ref{alg:score} explains our metric. 
    \fi
    
    \item \textbf{robustness,} where we check by how much the model performances degrade due to errors in the dataset.
    One core idea of \gls*{tdlgm} is that interleaving state and latent variables should facilitate robust prediction.
    That is, the generative model should still be able to perform given uncertainties in data.
    One previously discussed method of increasing robustness is to train a model on noisy data.
    \gls*{dlgm}, for example, struggled with recognition of unseen data without this \cite{dlgm}.
    We, therefore, want \gls*{tdlgm} to reconstruct unseen data without having it train on noisy data.
    Our tests prove that \gls*{tdlgm} can reconstruct without training on noisy data, validating the claim that \gls*{tdlgm} is robust.
    \if
    Another form of robustness is the ability of \gls*{tdlgm} to reconstruct data when parts of it are erroneous.
    That is, we want to reconstruct not only data riddled with noise but also data that contains pure inaccuracies.
    We have chosen to do this by replacing the last four values in any reconstruction task with values in the $[0,1]$ range.
    The last four values were chosen based on the assumption that more recent values should have a more significant impact on the next.
    \fi
\end{enumerate}

\begin{algorithm}[!h]
 \caption{Evaluation of future prediction.
 Models are scored by grouping their generated values into intervals.
 The algorithm then counts how many instances of a specific value are generated after another value.
 Therefore, distributions are formed over what values are usually generated when.
 The overlap between the true and generated distributions is then used as the score.
 It assumes that $T$ and $G$ contain a finite set of values corresponding to intervals, or so-called buckets of values.
 Scaling is also performed, accounting for the fact that $T$ and $G$ might have different amounts of values. 
 }
 \label{alg:score}
 \KwData{$T \leftarrow \text{true data}$ }
 \KwData{$G \leftarrow \text{generated data}$ }
 \KwData{$s \leftarrow \text{step forward in time}$}
 \KwData{$TM \leftarrow \text{Matrix initialized to zero}$}
 \KwData{$GM \leftarrow \text{Matrix initialized to zero}$}
 \For{$t_i \in T, \forall t_i, \text{where } t_{i+s}\in T$}{
    $TM_{t_i, t_{i+s}} \leftarrow TM_{t_i, t_{i+s}} + 1$
  }
  \For{$g_i \in G, \forall t_i, \text{where } g_{i+s}\in G$}{
    $GM_{g_i, g_{i+s}} \leftarrow GM_{g_i, g_{i+s}} + 1$
  }
  $GM \leftarrow \nicefrac{|T|}{|G|}\times GM$ Scale values in $GM$ based on amount of data points.
  \newline
  $score = \sum_i\sum_j\nicefrac{\min(GM_{i,j},TM_{i,j})}{TM_{i,j}} , \forall i,j \  TM_{i,j} \neq 0$
\end{algorithm}

\section{\uppercase{Results}}

We, as expected, found that \gls*{dlgm} performed best when it came to the reconstruction of data.
With \gls*{tdlgm} as a second, the basic \gls*{rnn} performs worst.
See Table \ref{table:recon} for a complete table of the different values that were tried and their mean squared error\footnote{
Readers may have noticed that the choice of modifier probability and variance is arbitrarily picked without providing exhaustive tests over the range. 
This is an explicit choice as we are interested in how the models compare against each other, not in how well the models fit the dataset.
}.
Here, it is evident that \gls*{dlgm} was almost unaffected by the noise, generating accurate values with close to the same variance for all cases.
\gls*{rnn} is also unaffected. 
However, this is because the best model, according to the mean squared metric, produces bogus results.
Figure \ref{fig:rnn_best_recon} shows a slice of the best reconstruction, and \ref{fig:rnn_worse_recon} shows a worse slice (according to this metric).
However, it is clear that the \textit{worse} model follows the pattern while the better scoring model does not.
Therefore, a filtering process is applied by performing a t-test to see if there is any significant difference between the reconstructed data given different levels of noise. 
This filtering assumes that the reconstruction is negatively affected to such a degree that it is statistically significant. 
All trained instances with a p-value above $0.7$ were excluded, leaving us with the new values we call Filtered \gls*{rnn} in Table \ref{table:recon}.
The high p-value is set to counteract the risk of confirmation bias.
We assume that the error should increase with the noise but do not want to filter out evidence proving the contrary.

Our new results make it evident that \gls*{rnn} is much more sensitive to noise as compared to both \gls*{tdlgm} and \gls*{dlgm}.
Hence, we conclude that \gls*{tdlgm} performs better than the regular \gls*{rnn} at value imputation. 
Figures \ref{fig:tdlgm_rec1}, \ref{fig:dlgm_rec1}, and \ref{fig:rnn_rec1} show the reconstruction without any noise for \gls*{tdlgm}, \gls*{dlgm}, and \gls*{rnn} respectively.
Reconstruction with noise can be seen in Figures \ref{fig:rnn_rec}, \ref{fig:dlgm_rec}, and \ref{fig:tdlgm_rec}.
\glspl*{dlgm} ability to perfectly reconstruct should not be interpreted as better than \gls*{tdlgm}.
The simplicity of the data means that \gls*{dlgm} can easily overfit all possible latent states. 
Besides, more or less perfect reconstruction is far from the goal and often unattainable, given a limited dataset.
It can then be better to have a reconstruction model such as \gls*{tdlgm} that creates close to similar, but not the exact values.
Future research should be done with higher dimensional and more complex temporal patterns to see how this changes performance.

\begin{figure}
    \centering
    \includegraphics[width=0.5\textwidth]{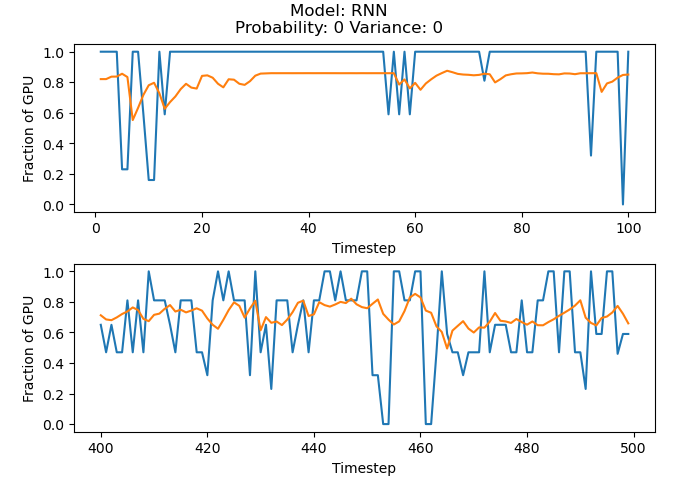}
    \caption{The best performing \gls*{rnn} model for reconstruction when using a naive \gls*{mse} scoring method. Blue are the true values and orange is the generated values. The Y-axis specifies what fraction of a GPU a task at a specific step requires, with a maximum of 1 and minimum of 0. It is evident that the reconstruction process performs poorly.}
    \label{fig:rnn_best_recon}
\end{figure}

\begin{figure}
    \centering
    \includegraphics[width=0.5\textwidth]{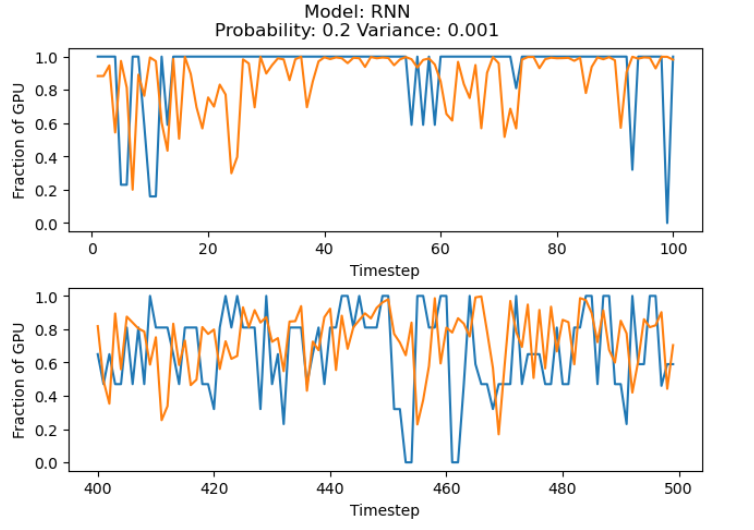}
    \caption{A model that performed worse than the best performing \gls*{rnn} when using a naive \gls*{mse} scoring method. Blue represents true values, and orange represents the generated values. The Y-axis specifies what fraction of a GPU a task at a specific step requires, with a max of 1 and a minimum of 0. It is evident that this reconstruction is more true to the actual test data, as compared to \ref{fig:rnn_best_recon}, despite the fact that it got a lower score. The probability and variance denote the introduction of error. With the probability denoting the likelihood of modifying a value and the variance specifying the magnitude of change according to a Gaussian. }
    \label{fig:rnn_worse_recon}
\end{figure}

\begin{figure}
\centering
    \begin{subfigure}[b]{0.5\textwidth}
        \includegraphics[width=\textwidth]{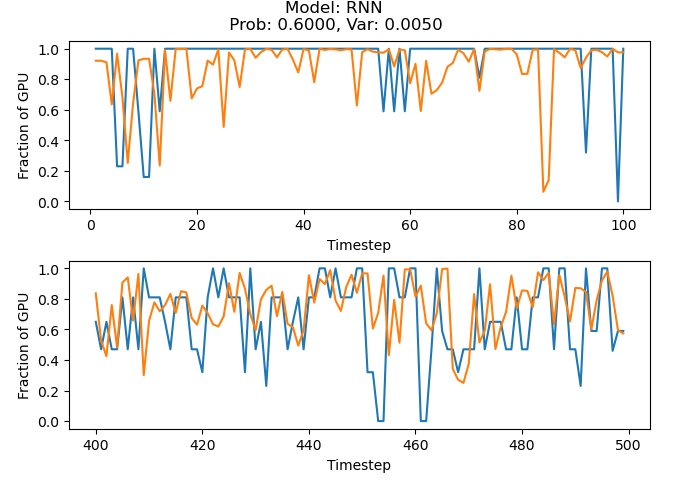}
        \caption{Probability of modifying value: 60\%, modifying through a Gaussian with variance set to $0.005$.}
        \label{rnn:sub1}
    \end{subfigure}
    \begin{subfigure}[b]{0.5\textwidth}
        \includegraphics[width=\textwidth]{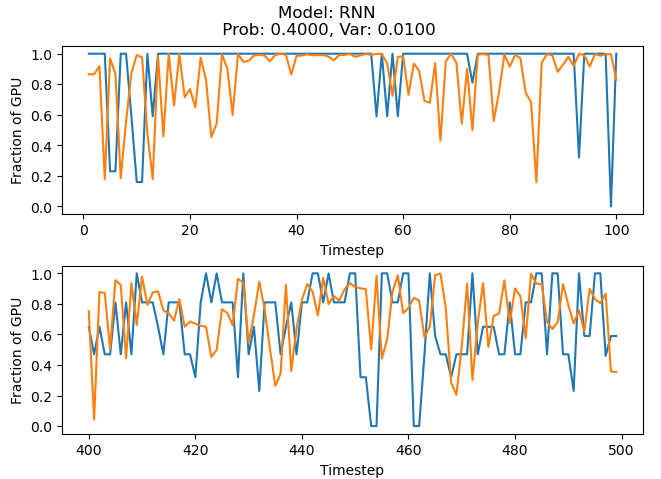}
        \caption{Probability of modifying value: 40\%, modifying through a Gaussian with variance set to $0.01$.}
        \label{rnn:sub2}
    \end{subfigure}
    \caption{\gls*{rnn} tasked with reconstructing the test dataset. The orange line is generated values, and the blue is true values.
    The X-axis is time.
    The Y-axis specifies what fraction of a GPU a task requires at a specific step, with a maximum of 1 and a minimum of 0.
    }
    \label{fig:rnn_rec}
\end{figure}

\begin{figure}
\centering
    \begin{subfigure}[b]{0.5\textwidth}
        \includegraphics[width=\textwidth]{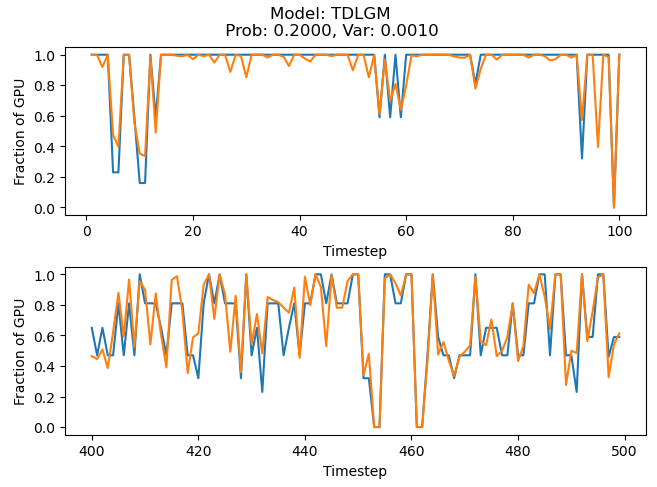}
        \caption{Probability of modifying value: 20\%, modifying through a Gaussian with variance set to $0.001$.}
        \label{tdlgm:sub1}
    \end{subfigure}
    \begin{subfigure}[b]{0.5\textwidth}
        \includegraphics[width=\textwidth]{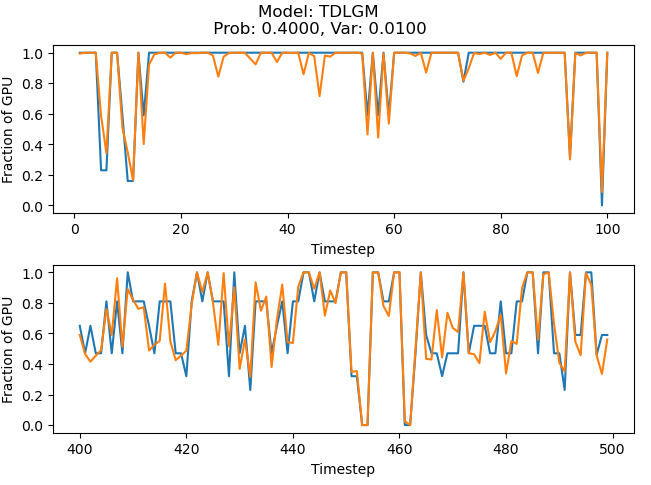}
        \caption{Probability of modifying value: 40\%, modifying through a Gaussian with variance set to $0.01$.}
        \label{tdlgm:sub2}
    \end{subfigure}
    \caption{\gls*{tdlgm} tasked with reconstructing the test dataset. The orange line is generated values, and the blue is true values.
    The X-axis is time.
    The Y-axis specifies what fraction of a GPU a task requires at a specific step, with a maximum of 1 and a minimum of 0.
    Subfigures \ref{tdlgm:sub1} and \ref{tdlgm:sub2} show the model's performance with different noise levels added
    }
    \label{fig:tdlgm_rec}
\end{figure}

\begin{figure}
\centering
    \begin{subfigure}[b]{0.5\textwidth}
        \includegraphics[width=\textwidth]{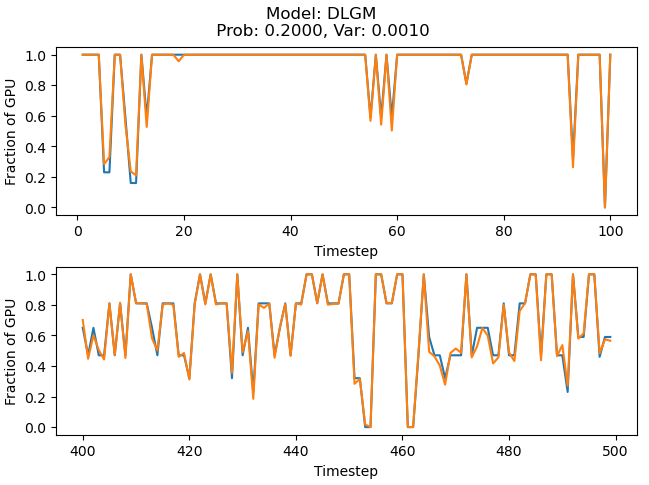}
        \caption{Probability of modifying value: 20\%, modifying through a Gaussian with variance set to $0.001$.}
        \label{dlgm:sub1}
    \end{subfigure}
    \begin{subfigure}[b]{0.5\textwidth}
        \includegraphics[width=\textwidth]{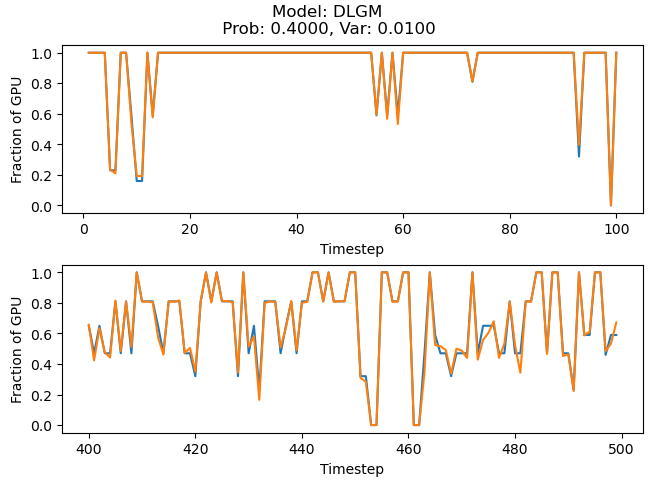}
        \caption{Probability of modifying value: 40\%, modifying through a Gaussian with variance set to $0.01$.}
        \label{dlgm:sub2}
    \end{subfigure}
    \caption{\gls*{dlgm} tasked with reconstructing the test dataset. The orange line is generated values, and the blue is true values.
    The X-axis is time.
    The Y-axis specifies what fraction of a GPU a task requires at a specific step, with a maximum of 1 and a minimum of 0.
    Subfigures \ref{dlgm:sub1} and \ref{dlgm:sub2} show the model's performance with different noise levels added
    }
    \label{fig:dlgm_rec}
\end{figure}

\begin{table}
\caption{Table showing how well the different models reconstructed the test data. Each data point had a percentual chance of being modified through a Gaussian sample. 
The variance was modified to see how each model reacted to different magnitudes of error.}
\label{table:recon} \centering
\begin{tabular}{|c|c|c|}
  \hline
  \multicolumn{3}{|c|}{ Probability: 0, Variance: 0 }\\
  \hline
  Model & Mean & Variance \\
  \hline
  \gls*{tdlgm} & -0.0014 & 0.0122 \\
  \hline
  \gls*{dlgm} &  -0.0036 & 0.0006 \\
  \hline
  \gls*{rnn} & -0.0344 &  0.0736 \\
  \hline
  Filtered \gls*{rnn} &  0.0124 &  0.1038 \\
  \hline
  \multicolumn{3}{|c|}{ Probability: 0.2, Variance: 0.001 }\\
  \hline
  Model & Mean & Variance \\
  \hline
  \gls*{tdlgm} & -0.0015 & 0.0122 \\
  \hline
  \gls*{dlgm} & -0.0036 & 0.0006 \\
  \hline
  \gls*{rnn} & -0.0344 & 0.0736 \\
  \hline
  Filtered \gls*{rnn} &  -0.0104 &  0.1027 \\
  \hline
  \multicolumn{3}{|c|}{ Probability: 0.6, Variance: 0.005 }\\
  \hline
  Model & Mean & Variance \\
  \hline
  \gls*{tdlgm} & -0.0096 & 0.0118 \\
  \hline
  \gls*{dlgm} &  -0.0041 & 0.0006 \\
  \hline
  \gls*{rnn} & -0.0344 &  0.0736 \\
  \hline
  Filtered \gls*{rnn} &  0.0130 &  0.1032 \\
  \hline
  \multicolumn{3}{|c|}{ Probability: 0.4, Variance: 0.01 }\\
  \hline
  Model & Mean & Variance \\
  \hline
  \gls*{tdlgm} & -0.0006 & 0.0114 \\
  \hline
  \gls*{dlgm} &  -0.0040 & 0.0006 \\
  \hline
  \gls*{rnn} & -0.0344 &  0.0736 \\
  \hline
  Filtered \gls*{rnn} &  -0.0089 &  0.1149 \\
  \hline
  \multicolumn{3}{|c|}{ Probability: 0.8, Variance: 0.05 }\\
  \hline
  Model & Mean & Variance \\
  \hline
  \gls*{tdlgm} & 0.0008 & 0.0118 \\
  \hline
  \gls*{dlgm} &  -0.0038 & 0.0007 \\
  \hline
  \gls*{rnn} & -0.0344 &  0.0736 \\
  \hline
  Filtered \gls*{rnn} &  -0.0014 &  0.1134 \\
  \hline

\end{tabular}
\end{table}

\begin{figure}
    \centering
     \includegraphics[width=0.5\textwidth]{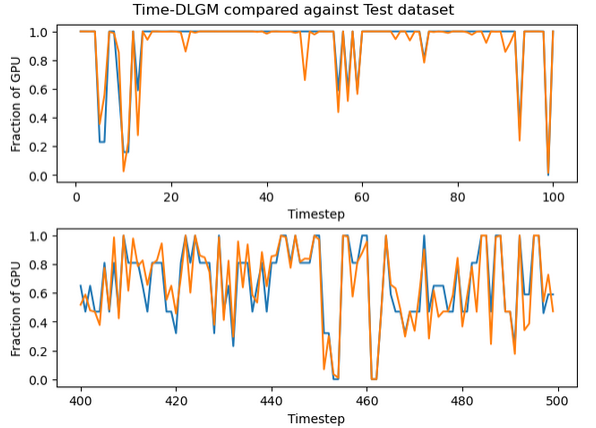}
    \caption{\gls*{tdlgm} tasked with reconstructing the test dataset. Orange is generated values and blue is true values. The X-axis is time. The Y-axis specifies what fraction of a GPU a task at a specific step requires with a max of $1$ and $0$ as minimum. It is evident from the figure that \gls*{tdlgm} can reconstruct the data.}
    \label{fig:tdlgm_rec1}
\end{figure}

\begin{figure}
    \centering
    \includegraphics[width=0.5\textwidth]{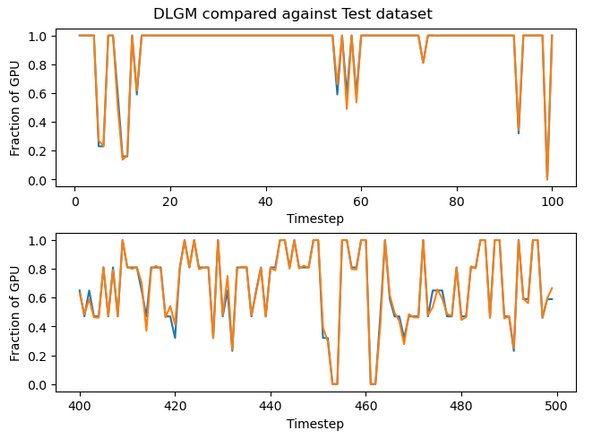}
    \caption{\gls*{dlgm} compared against the test dataset. Orange is the generated values, and blue is the true values. The X-axis is the steps in an unspecified time unit, where one generation is performed each step. The Y-axis specifies what fraction of a GPU a task at a specific step requires with a max of $1$ and $0$ as minimum. \gls*{dlgm} can perfectly reconstruct values in many instances, as evident by the high overlap.}
    \label{fig:dlgm_rec1}
\end{figure}

\begin{figure}
\centering
        \includegraphics[width=0.5\textwidth]{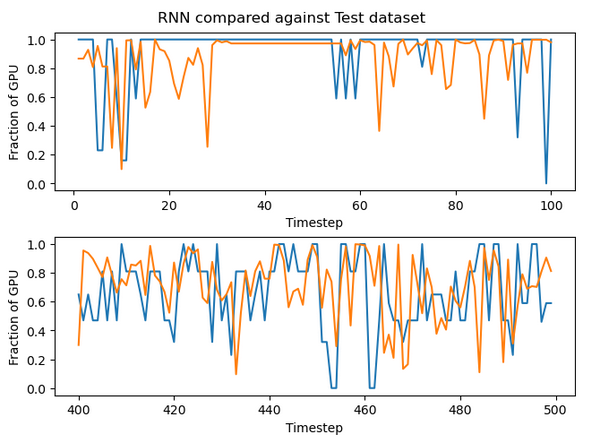}
        \caption{\gls*{rnn} tasked with reconstructing the test dataset. Orange is generated values and blue is true values. The X-axis is time. The Y-axis specifies what fraction of a GPU a task at a specific step requires with a max of $1$ and $0$ as minimum. It is evident from the figure that \gls*{rnn} can reconstruct data, although much worse than the other two models.}
        \label{fig:rnn_rec1}
\end{figure}

Our multiple time-step prediction is based on similar distributions, not identical values.
This means that we do not care about individual values but about the distribution of future values. 
The previously discussed Algorithm \ref{alg:score} is therefore used as our metric.
Scoring is done on a dataset of future generations.
It was generated by giving each model a sequence of values, which was then used to create future values see Table \ref{table:future} for results.
\gls*{rnn} perform better when there are a few amounts of steps in the future.
However, it quickly deteriorates when generating longer sequences.
While \gls*{tdlgm} is stable with close to the same score no matter how long or short the generated sequence.
\gls*{dlgm} also exhibit this stable behavior but with a score worse than \gls*{tdlgm}.
We conclude based on this that \gls*{tdlgm} provides a more consistent future generation compared to \gls*{rnn}.

\begin{table}
\caption{Table showing how well the different models generated future values. 
A higher score correlates with a larger overlap between the true and generated values.
Scoring is done with Algorithm \ref{alg:score}.
\textit{Steps} defines the number of consecutive digits generated for each test.
For example, a sequence length of 30 means that a state was fed, which was then used to create the following 30 values.
We can conclude from this that \gls*{tdlgm} was best at generating future values that look similar to the actual distribution, followed by \gls*{dlgm} and then \gls*{rnn}. 
}
\label{table:future} \centering
\begin{tabular}{|c|c|c|c|}
\hline
  Steps & \gls*{tdlgm} & \gls*{rnn} & \gls*{dlgm} \\
  \hline
  2 & 62.47 & 58.48 & 59.87 \\
  \hline
  5 & 63.47 & 58.48 & 59.67 \\
  \hline
  8 & 63.21 & 58.41 & 60.32 \\
  \hline
  10 & 63.59 & 58.04 &  59.67 \\
  \hline
  15 & 63.67 & 57.79 & 60.11 \\
  \hline
  20 & 63.54 & 58.01 & 60.36 \\
  \hline
  25 & 63.31 & 57.80 &  59.89 \\
  \hline
  30 & 63.31 & 58.01 & 60.33 \\
  \hline
\end{tabular}
\end{table}

We have now evaluated the model's ability to create future values and perform inference and will therefore discuss the robustness of \gls*{tdlgm}.
The previous models were trained by combining true values with noise, a common practice to improve a model's robustness.
If \gls*{tdlgm} shows a good performance in this training scenario, then it is a testament to its robust properties.
We evaluated this by training \gls*{tdlgm} with the pure, unmodified training data and then performed reconstructions. 
Reconstructions were, as before, performed with varying layers of noise.
The difference now is that we are interested in robustness.
This means that we want to see how \gls*{tdlgm} performance is changed by different magnitudes of noise.
Our results can be seen in Table \ref{table:tdlgm_robust}.
Here, it is evident that the reconstruction was worse without the added noise.
Although it was still significantly better than that of \gls*{rnn}, which can be seen in Table \ref{table:recon}.
We conclude based on this that \gls*{tdlgm} contain inherent properties, giving it the robustness to work on values outside of the training set without the need for any standard robustness-inducing techniques.
Figures \ref{fig:rob_1} show the reconstruction for two cases of additive noise.

\begin{table}
\caption{Table showing how well \gls*{tdlgm} managed to reconstruct unseen values based on different amounts of noise.
Each data point in the test dataset was modified according to the Gaussian $\mathcal{N}(0,\sigma^2)$ where $\sigma^2$ is varied variance.
The output was clamped to the range $[0,1]$. If any data point was given a value over $1$ or under $0$, it was moved to its corresponding minimum or maximum.
\gls*{tdlgm} was trained purely on the unmodified training data.
This was done to evaluate how robust \gls*{tdlgm} is to unknown data.
}
\label{table:tdlgm_robust} \centering
\begin{tabular}{|c|c|c|}
\hline
  $\sigma^2$ & Mean & Variance  \\
  \hline
  0.0053 & -0.01968 & 0.016377 \\
  \hline 
  0.0059 & -0.017715 & 0.017023 \\
  \hline
  0.0067 & -0.020703 & 0.016572 \\
  \hline
  0.0077 & 0.016572 & 0.016885 \\
  \hline
  0.0091 & -0.018863 & 0.016863 \\
  \hline
  0.0111 & -0.019976 & 0.016314 \\
  \hline
  0.0143 & -0.016961 & 0.017189 \\
  \hline
  0.0200 & -0.019446 & 0.017566 \\
  \hline
  0.0333 & -0.026147 & 0.017805 \\
  \hline
  0.100 & -0.039585 & 0.027555 \\ 
  \hline
\end{tabular}
\end{table}

\begin{figure}
\centering
    \begin{subfigure}[b]{0.5\textwidth}
        \includegraphics[width=\textwidth]{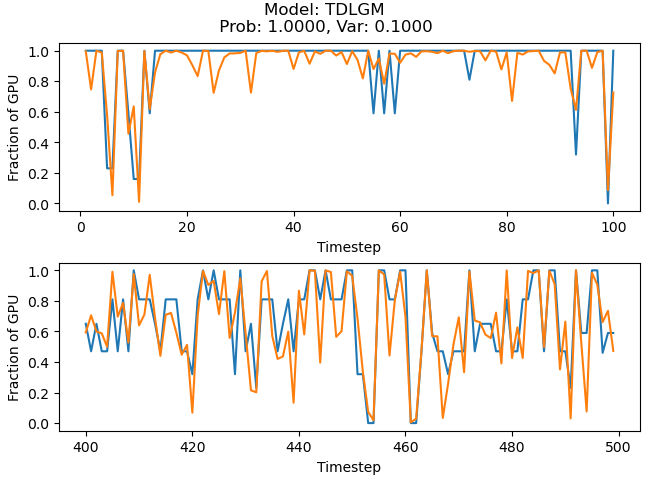}
        \caption{Probability of modifying value: 100\%, modifying through a Gaussian with variance set to $0.1$.}
        \label{tdlgm:sub3}
    \end{subfigure}
    \begin{subfigure}[b]{0.5\textwidth}
        \includegraphics[width=\textwidth]{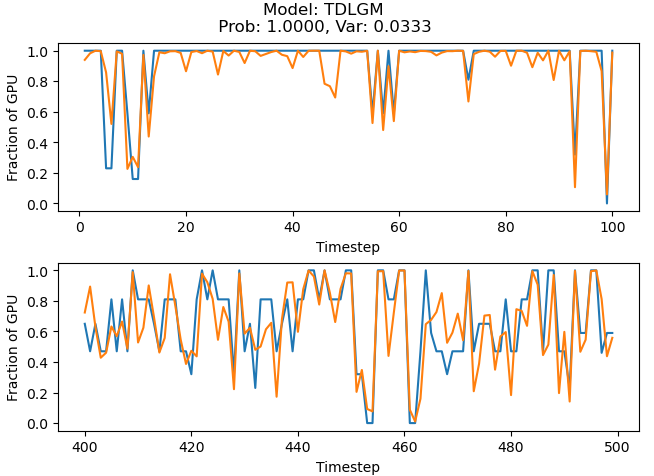}
        \caption{Probability of modifying value: 100\%, modifying through a Gaussian with variance set to $0.0333$.}
        \label{tdlgm:sub4}
    \end{subfigure}
    \caption{\gls*{tdlgm} tasked with reconstructing the test dataset. The orange line is generated values, and the blue is true values.
    The X-axis is time.
    The Y-axis specifies what fraction of a GPU a task requires at a specific step, with a maximum of 1 and a minimum of 0.
    Subfigures \ref{tdlgm:sub3} and \ref{tdlgm:sub4} show the model's performance with different noise levels added
    }
    \label{fig:rob_1}
\end{figure}

\section{\uppercase{Conclusion}}


We have in this paper proposed a new model called \gls*{tdlgm}, based on the ideas of a recognition-generator structure with state.
\gls*{tdlgm} differ from other models commonly seen in the literature by using two recognition models, one for state and one for latent variables. 
Both state and latent variables are combined in an interleaving structure, which allows for the learning of complex temporal relations. 
Furthermore, the state recognition model and the regularization we derive from it are, to our knowledge, a novel method of inference.
Our experiments on the Alibaba trace dataset \cite{alibaba} show that \gls*{tdlgm} is a promising model. 
Showing good performance with regard to imputation, the generation of new values, and robustness.

However, there are also some points for future improvements. 
One such point is the approximation as seen in Equation \eqref{mseloss}.
It was derived to compensate for the fact that a state distribution is unknown.
Future work can explore this unknown factor to determine whether better approximations can be performed or if none are needed.
We also found it more challenging to train \gls*{tdlgm} than \gls*{dlgm}, with many models getting stuck in local minimums.
We believe this ties into our approximation. 
Another avenue of future research is to make a more exhaustive comparison against other models for time series data, such as \gls*{vrnn} \cite{vrnn}.

The code for \gls*{tdlgm} can be found \href{https://git.cs.kau.se/johaanto/tdlgm}{here}.

\bibliographystyle{apalike}
{\small
\bibliography{example}}



\end{document}